\documentclass[10pt,twocolumn,letterpaper]{article}

\usepackage{iccv}
\usepackage{times}
\usepackage{epsfig}
\usepackage{graphicx}
\usepackage{amsmath}
\usepackage{amssymb}

\usepackage[accsupp]{axessibility}

\usepackage{multirow}
\usepackage{ntheorem}
\usepackage{capt-of}
\usepackage{adjustbox}
\usepackage{array}
\usepackage{booktabs}
\theoremstyle{break}
\usepackage{bm}
\newtheorem{definition}{Definition}
\usepackage[pagebackref=true,breaklinks=true,letterpaper=true,colorlinks,bookmarks=false]{hyperref}

\iccvfinalcopy 

\def\iccvPaperID{4148} 

\ificcvfinal\fi

\begin{document}

\title{Progressive Seed Generation Auto-encoder

for Unsupervised Point Cloud Learning}

\author{Juyoung Yang \qquad Pyunghwan Ahn \qquad Doyeon Kim \qquad Haeil Lee \qquad Junmo Kim\\
School of Electrical Engineering, KAIST, South Korea\\
{\tt\small \{yjy6711, p.ahn, doyeon\_kim, haeil.lee, junmo.kim\}@kaist.ac.kr}
}

\maketitle
\ificcvfinal\thispagestyle{empty}\fi

\begin{abstract}
With the development of 3D scanning technologies, 3D vision tasks have become a popular research area.
Owing to the large amount of data acquired by sensors, unsupervised learning is essential for understanding and utilizing point clouds without an expensive annotation process.
In this paper, we propose a novel framework and an effective auto-encoder architecture named ``PSG-Net'' for reconstruction-based learning of point clouds. Unlike existing studies that used fixed or random 2D points, our framework generates input-dependent point-wise features for the latent point set.
PSG-Net uses the encoded input to produce point-wise features through the seed generation module and extracts richer features in multiple stages with gradually increasing resolution by applying the seed feature propagation module progressively. We prove the effectiveness of PSG-Net experimentally; PSG-Net shows state-of-the-art performances in point cloud reconstruction and unsupervised classification, and achieves comparable performance to counterpart methods in supervised completion.
\end{abstract}
\vspace{-0.5cm}
\section{Introduction}
\label{sec:sec1}
Deep neural networks, especially convolutional neural networks (CNNs), have achieved success in the performance of various computer vision tasks~\cite{imagenet,fcn,ssd}. Recently, research centered on 2D space has been expanded to 3D space following the development of 3D scanning techniques. 3D sensors such as LiDAR and RGB-D cameras acquire data in form of point clouds, so it is essential to effectively recognize this type of data in robotics and autonomous driving applications.
Point clouds lie in 3D space, so they incur significantly higher labeling costs for a specific vision task when compared to 2D data. Therefore, the need for effective unsupervised learning techniques for point clouds is highly emphasized.

\begin{figure}[!ht]
\begin{center}
\includegraphics[trim=6cm 17cm 6.5cm 17cm,clip,width=\linewidth]{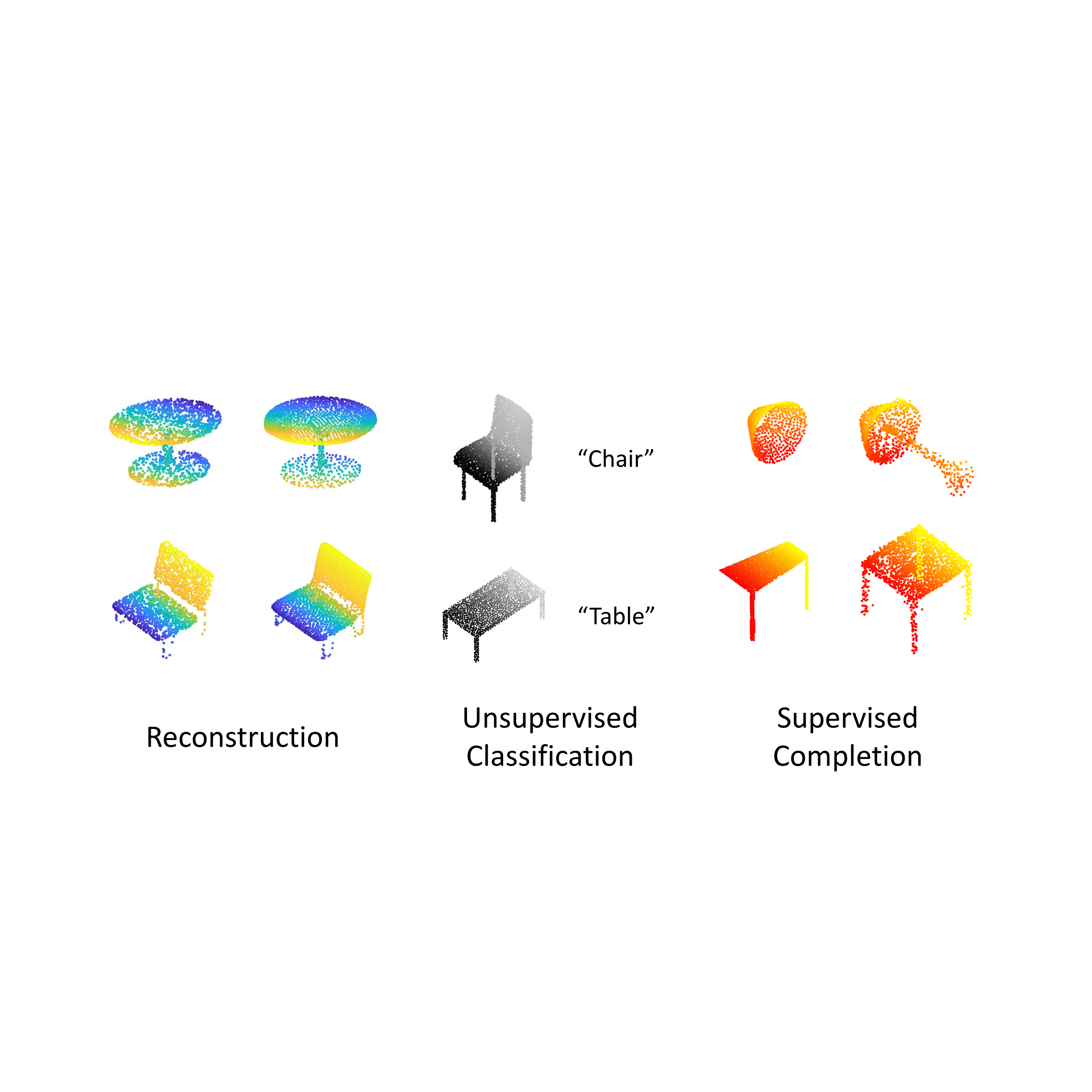}
\end{center}
\caption{Examples of the 3D point cloud reconstruction, unsupervised classification, point cloud completion results with PSG-Net. Our network successfully performs each task from the input point cloud.}
\label{fig:fig0}
\vspace{-0.3cm}
\end{figure}

For unsupervised learning, auto-encoder structures based on CNNs have been widely used to deal with 2D image data. Although CNNs show superior ability in learning general features from 2D images, it is difficult to apply CNNs to point clouds because of the irregularity of the data format.
Thus, network architectures that are specifically designed for point cloud recognition must be used as the encoder of the auto-encoder.
In previous studies~\cite{atlasnet, 3dpointcapsulenet, sanet}, point cloud classification networks such as PointNet~\cite{pointnet} and PointNet++~\cite{pointnet++} have been used as encoders, and some other works have utilized graph layers~\cite{foldingnet}. Encoders extract a global feature representation called a codeword vector, which becomes the input to the decoder.

For the decoder architecture, a popular approach is the concept of ``folding'' a 2D plane into a 3D object surface because the number of the output point clouds may not be determined. During the folding operation, the latent point set sampled from the 2D plane is transformed into 3D points, thus achieving point cloud generation. This approach was first proposed in FoldingNet~\cite{foldingnet} and has been mathematically established in several studies~\cite{atlasnet, 3dpointcapsulenet}.
In FoldingNet, a pre-defined set of fixed grid points in 2D space was used as the latent point set to the decoder along with the output of the encoder. Then, the folding operation was applied to transform this 2D point set into 3D points. In AtlasNet~\cite{atlasnet} and 3D Point Capsule Network (3D-PointCapsNet)~\cite{3dpointcapsulenet}, multi-patch approaches were used for point cloud reconstruction. These works used randomly sampled points from the uniform distribution in a fixed area of the 2D plane. Throughout this paper, the latent point set will be referred to as \textit{seed}.

One limitation of these methods is that the output point cloud is generated from an arbitrary 2D plane. Fixed grid points~\cite{foldingnet} or randomly sampled points~\cite{atlasnet, 3dpointcapsulenet} were used as inputs to the decoder. Considering that an arbitrary 2D plane might not have enough capacity to model a complex 3D surface, AtlasNet and 3D-PointCapsNet utilize multiple patches to improve performance. However, the number of decoders increases with the number of patches, which leads to significant computational costs. Therefore, it is necessary to fundamentally change the sampling process in the 2D plane, rather than simply use more patches.

In this paper, we propose a novel framework for reconstruction-based learning. The main idea is to generate the seed 
from the function of the input point cloud. Previous methods used fixed or randomly sampled 2D points as seeds, which can be a substantial constraint to the decoder. We revisit the problem to explain why the generated seed helps our decoder to generate various 3D shapes. We implement our framework by ``PSG-Net'', which is far more effective and superior to existing methods. The seed is generated in multiple stages through the seed generation module. Then, the seed feature propagation module processes the generated seed and codeword vector to produce the output shape. In addition, we introduce a progressive approach to enrich the information of the feature for the output point cloud. This is achieved by generating a gradually increasing resolution of the seed and interpolated feature maps. The result from the last seed feature propagation module is transformed into 3D point cloud coordinates through the point generation layers.

Our network achieves performances comparable to the state-of-the-art methods in various unsupervised tasks such as point cloud reconstruction and unsupervised classification, and achieves the best performance among counterpart methods in supervised point cloud completion. Furthermore, our method can be used in combination with other existing methods such as the multi-patch approach, which is expected to result in enhanced performance. The main contributions of this study are as follows.
\begin{enumerate}
    \item We propose a novel framework for reconstruction-based learning of point clouds that uses the generated input-dependent point-wise features as seed, instead of using simple 2D points. To implement this framework, we incorporate seed generation module (SGM) and seed feature propagation module (SFPM) in an efficient auto-encoder architecture called PSG-Net.
    \item We analyze two proposed modules and demonstrate the superiority of our model by experimentally proving the analysis.
    \item We show that the performance of PSG-Net is comparable to state-of-the-art methods in various 3D tasks such as point cloud reconstruction, unsupervised classification and supervised point cloud completion.
\end{enumerate}

\section{Related work}

There has been much research on 3D computer vision using point clouds in recent years~\cite{pointnet, pointconv, dgcnn, pointcnn}. We present previous studies related to our method in this section. Because point cloud completion can be seen as a task expanded from point cloud reconstruction, a method for one of these tasks can often be used for another, as in this paper. Given that both tasks have been performed in this study as well, we introduce research on these two tasks.


\textbf{Point cloud reconstruction.}
The first auto-encoder was a simple network based on PointNet introduced in L-GAN~\cite{lgan}. L-GAN aims to generate the point cloud with generative adversarial network (GAN) and acts as a baseline network for later studies. Subsequently, FoldingNet~\cite{foldingnet} performs point cloud reconstruction by learning 2D-to-3D mapping under the intuition that a 2D plane can be transformed into a 3D surface through certain operations. The authors call this a ``folding'' operation and define it as the concatenation of replicated codeword vectors to low-dimensional grid points, followed by a point-wise multi-layer perceptron (MLP). FoldingNet consists of a graph-based encoder and a folding-based decoder, and uses a fixed 2D grid as the sampled points of a 2D plane.
AtlasNet~\cite{atlasnet} proposed to use multiple 2D patches instead of a single 2D plane as in FoldingNet. In this architecture, 2D points were randomly sampled from a uniform distribution inside the unit square, so that the network can better learn the shape of the objects.
3D-PointCapsNet~\cite{3dpointcapsulenet} extracts latent capsules by applying a dynamic routing system and creates a point cloud from them. There is also a study that performs reconstruction to learn local descriptors. PPF-FoldNet~\cite{ppffoldnet} converts a point cloud into a point pair feature (PPF) representation and performs feature reconstruction using folding operations to learn local descriptors on a point cloud.

\begin{figure*}[!ht]
\begin{center}
\includegraphics[width=11cm]{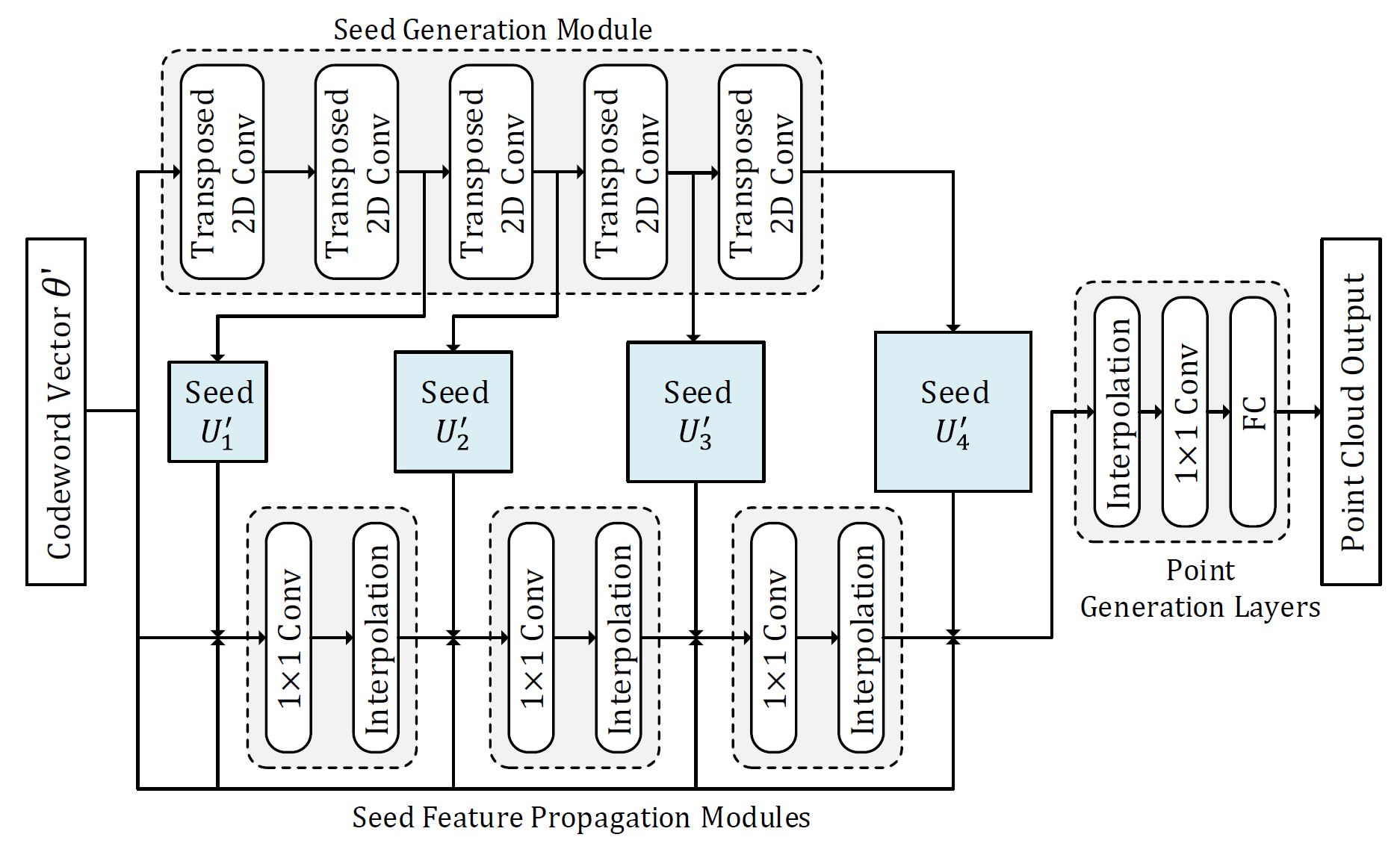}
\end{center}
\vspace{-0.1cm}
\caption{Overview of our framework. PSG-Net consists of three parts : seed generation module (SGM), seed feature propagation module (SFPM), and point generation layers. Several parameters such as the number of layers that make up the SGM or the number of SFPMs can be changed depending on the input conditions. Each convolution and fully connected layer is followed by batch normalization and ReLU layers (omitted from the figure).}
\label{fig:fig1}
\vspace{-0.1cm}
\end{figure*}

\textbf{Point cloud completion.}
Point cloud completion is a rapidly growing research area derived from point cloud reconstruction. PCN~\cite{PCN} was the first method to perform point cloud completion with a leading-based approach to a point cloud instead of a voxel. PCN uses an extended version of PointNet as an encoder and a decoder that combines the fully connected decoder and the folding-based decoder to generate a dense output point cloud. In ~\cite{topnet}, a tree-structured decoder called TopNet was proposed to create an arbitrary grouping of points. Because the global feature extracted from the incomplete point cloud may result in loss of information about structural details, SA-Net~\cite{sanet} uses a skip-attention mechanism for utilizing hierarchical information obtained from the PointNet++ encoder in the folding-based decoder. RL-GAN-Net~\cite{rlgannet} and Render4Completion~\cite{render4completion} complete the point cloud more realistically by utilizing reinforcement learning and GAN. Recently, GRNet~\cite{grnet} achieved high performance by combining a voxel-based approach and a point cloud-based approach. There have also been studies focusing on missing parts by separating the known and missing parts~\cite{detailpreserved, pfnet}.

\section{Method}
In this section, we introduce our framework and PSG-Net architecture. 
First, we formulate the problem and explain the model architecture. Then, we describe the analysis, including a comparison with existing methods.

\subsection{Formulation and Notation}
\label{sec:sec_3_1}
We follow the same approximation as those presented in previous studies~\cite{foldingnet, atlasnet, 3dpointcapsulenet}. We consider that in the point cloud $P = \{p_{i} \in \mathbb{R}^{3}\}$ , every point $p$ comes from the surface $S$ of a 3D shape. Because $S$ is 2-manifold, there exists an open set $U \subset \mathbb{R}^{2}$ and an open set $W$, that satisfies $p \in W \subset \mathbb{R}^{3}$, such that $S \cap W$ is homeomorphic to U. The mapping $C: S \cap W \rightarrow U$ is called a chart, and its inverse $\Psi \equiv C^{-1}: U \rightarrow S \cap W$ is called a parameterization. Based on these notations, the problem is defined as follows:

\begin{definition}[Problem]
The point cloud reconstruction problem is defined as learning to generate 3D surface $S$ by finding the function $\Psi$ that satisfies $\Psi(U, \theta) = P' \approx P$, where $\theta$ is a codeword vector.
\end{definition}

In practice, we use a discrete set $U' = \{u_{j} \in \mathbb{R}^{2}\}$ instead of an open set $U$ because the input point cloud $P$ is a discrete sampled subset of $S$. Because previous studies have already formulated the problem in detail and proved the theorems needed to support this type of method, we refer the reader to \cite{foldingnet, atlasnet, 3dpointcapsulenet} for further information. Additionally, we use the term ``seed'' for $u$ in this study. Thus, it can be said that $p'_{j}$ from $P' = \{p'_{j} \in \mathbb{R}^{3}\}$ is generated from the corresponding seed $u_{j}$.

\begin{definition}[Seed]
We call the element $u$ of the discrete set $U'$ a \textbf{``seed''}, which generates the corresponding output point.
\end{definition}

\subsection{Model Architecture}
The input to our model is an $N \times d$ point cloud, where $d$ is basically 3 as each point cloud is composed of 3D coordinates $(x, y, z)$. Considering that our main focus is on the decoder, we utilize the existing feature extractor, which is commonly used in 3D tasks as the encoder. For the reconstruction task, we adopted PointNet++~\cite{pointnet++} as our encoder. PointNet++ consists of three set abstraction modules, each of which consists of sampling, grouping, and mini-PointNet layers. This encoder structure helps capture the hierarchical features and finally extracts a global feature vector $\theta'$ as the codeword vector. The encoder can be replaced by any point cloud processing architecture according to its relevance to the target task.

The proposed decoder contains three different parts: seed generation module (SGM), seed feature propagation module (SFPM), and point generation layers. We explain each module in detail in the following paragraphs. The overall architecture is shown in Figure~\ref{fig:fig1}.


\textbf{Seed generation module} takes $\theta'$ as input and generates 2D feature maps from a transposed 2D convolution (Conv) - batch normalization (BN)~\cite{bn} - ReLU layer sequence. We use the feature map $f_{l}$ from the $l=\{1, 2, \cdots ,L\} \,$th layer sequence as seed set $U'_{l} = \{u'_{il}\}$ for the SGM. Given that the feature map $f_{l}$ has a spatial size of $[h_{l}, w_{l}]$, the number of $u'$ in $U'_{l}$ is $N'_{l} = (h_{l} \times w_{l})$. $U'_{L}$ from the final layer of the SGM has $N_L$ seeds, which is equal to or close to the number of input points $N$. 

\textbf{Seed feature propagation module} takes three inputs: seed generated from the SGM, codeword vector, and intermediate feature map produced by the previous SFPM. As input to the $k=\{1, 2, \cdots ,K\} \,$th SFPM, the codeword vector ($\theta'$) is replicated $N'_{L-(K-k+1)}$ times and then concatenated with $U'_{L-(K-k+1)}$ and the output of the $(k-1)\,$th SFPM in the channel dimension. For the $1\,$st SFPM, we use $\theta'$ and $U'_{1}$ to create the input of the $1\,$st SFPM. 
The $k\,$th SFPM is made of a $1\times1$ Conv - BN - ReLU layer sequence and an interpolation layer. After the input passes through the layer sequence, the $k\,$th seed-wise feature with the channel dimension $|\theta'|/2$ is created. We interpolate the feature for the $(k+1)\,$th SFPM. This interpolation layer ensures that the output of the $k\,$th SFPM $V_{k} = \{v_{ik}\}$ has the same number of seeds as those of $U'_{L-K+k}$. For the interpolation layer, bilinear interpolation is adopted for efficiency.

Point generation layers take $\theta'$, $U'_{L}$, and $V_{K}$ to generate the output point cloud. The point generation layers are similar to the SFPM but have slightly different components to output a point cloud. Because the size of the generated point-wise feature may be different from the desired output size, we add the interpolation layer to match this. Then, the feature map is flattened and passed through a fully connected (FC) layer to produce an $M\times3$ point cloud as the output.

As in other studies~\cite{foldingnet, atlasnet, 3dpointcapsulenet}, we use the discrete chamfer distance as the loss function for training. The loss function is as follows:

\begin{multline}
  L_{CD}(P,P') = \cfrac{1}{|P|}\sum_{p \in P} \min_{p' \in P'} \lVert \mathbf{p - p'} \rVert_{2} +\\
    \cfrac{1}{|P'|}\sum_{p' \in P'} \min_{p \in P} \lVert \mathbf{p' - p} \rVert_{2}.
\end{multline}

For specific implementation details about our decoder, please refer to Supplementary Material Section 1.


%

\subsection{Analysis}
\label{sec:sec_3_3}
We introduced the formulated form of the problem in Section~\ref{sec:sec_3_1}. Now, we analyze our method and its superiority by revisiting the problem. If we represent the encoder as a function $E$, then $\theta$ can be written as $E(P)$. Therefore, the problem of point cloud reconstruction is to find $\Psi$ that satisfies $\Psi(U', E(P)) = P' \approx P$. Now, we can rewrite Definition 1 as follows:

\begin{definition}
The problem of point cloud reconstruction is defined as finding the function $\Psi$, where $\Psi(U', E(P))$ forms an identity function $I_{P}$.
\label{def:def3}
\end{definition}

The fundamental difference between our framework and the previous methods lies in generating $U'$. So far, $U'$ has been set to a fixed grid or randomly generated in the domain $]\,0, 1[\,^{2}: \mathcal{U}(0,1)$. That is, $u$ is an arbitrary 2D coordinate $(x, y)$ independent of $p$. However, the presence of another independent variable $U'$ may make it difficult to optimize $\Psi(U', E(P))$ to form $I_{P}$. We ease this optimization issue by setting $U'$ as a function of $P$. Because our decoder generates $U'$ from $\theta$ through the function $G$ that represents the SGM, $\Psi(U', E(P))$ can be expressed as $\Psi(G(\theta), E(P)) = \Psi(G(E(P)), E(P))$. Given that $\Psi(G(E(P)), E(P))$ can be rearranged into $\Psi'(P)$, the problem becomes easier to solve than it had been previously:

\begin{table*}[!ht]
  \begin{center}
  \centering
  \begin{tabular}{ ccccccc }
    (a) &
    \begin{minipage}{.12\linewidth}
      \includegraphics[trim=2cm 0.5cm 2cm 0.5cm,clip,width=\linewidth]{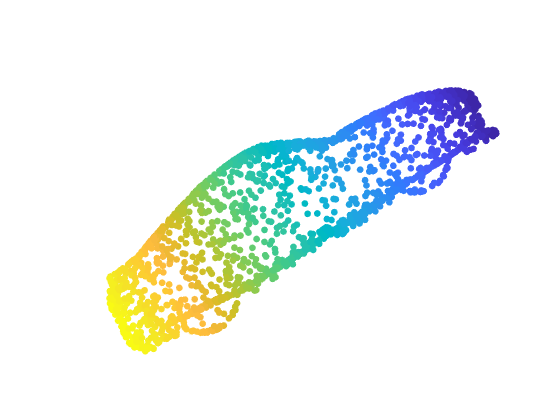}
    \end{minipage}
    & 
    \begin{minipage}{.12\linewidth}
      \includegraphics[trim=2cm 0.5cm 2cm 0.5cm,clip,width=\linewidth]{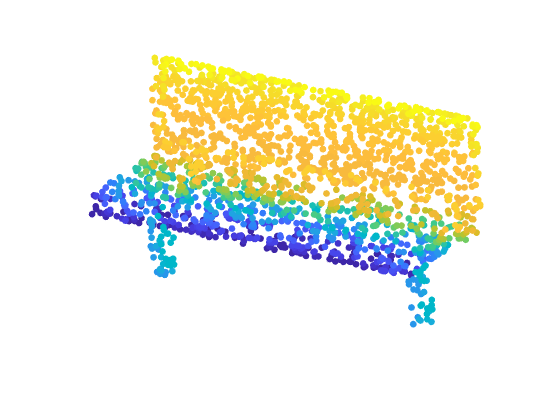}
    \end{minipage}
    &
    \begin{minipage}{.12\linewidth}
      \includegraphics[trim=2cm 0.5cm 2cm 0.5cm,clip,width=\linewidth]{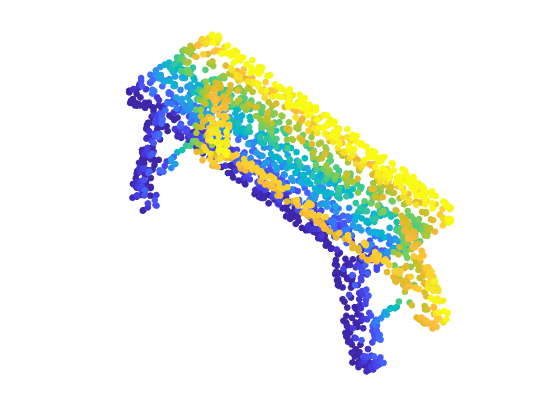}
    \end{minipage}
    &
    \begin{minipage}{.12\linewidth}
      \includegraphics[trim=2cm 0.5cm 2cm 0.5cm,clip,width=\linewidth]{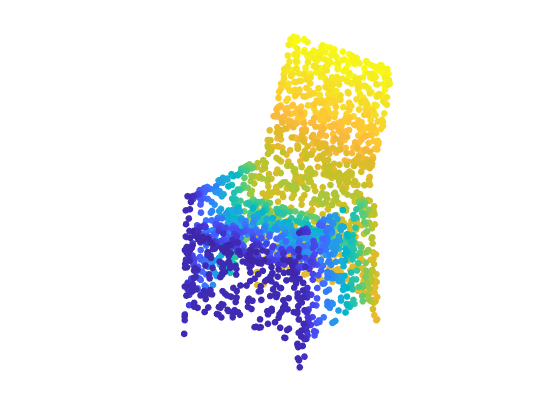}
    \end{minipage}
    & 
    \begin{minipage}{.12\linewidth}
      \includegraphics[trim=2cm 0.5cm 2cm 0.5cm,clip,width=\linewidth]{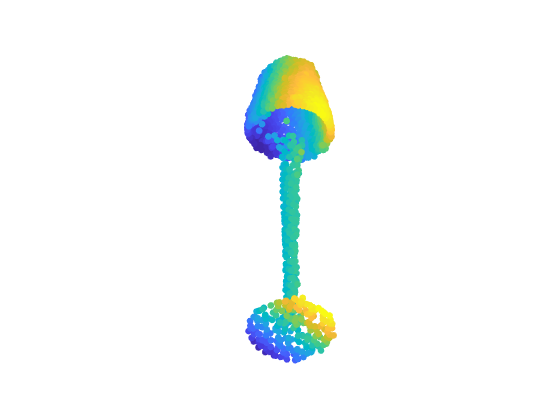}
    \end{minipage}
    &
    \begin{minipage}{.12\linewidth}
      \includegraphics[trim=2cm 0.5cm 2cm 0.5cm,clip,width=\linewidth]{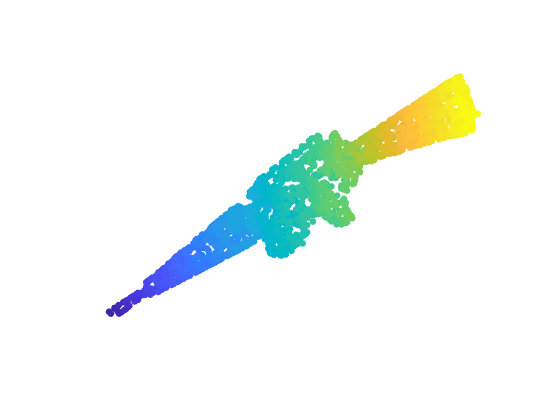}
    \end{minipage}
    \\ 
    (b) &
    \begin{minipage}{.12\linewidth}
      \includegraphics[trim=2cm 0.5cm 2cm 0.5cm,clip,width=\linewidth]{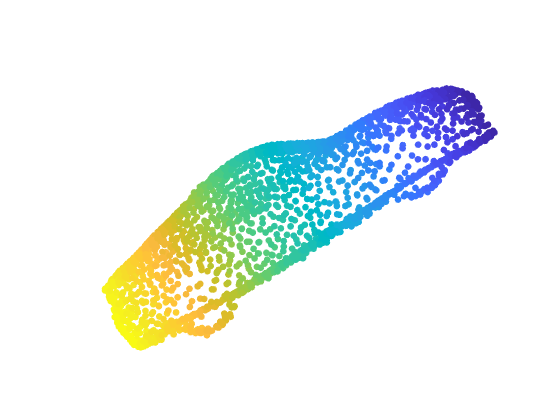}
    \end{minipage}
    & 
    \begin{minipage}{.12\linewidth}
      \includegraphics[trim=2cm 0.5cm 2cm 0.5cm,clip,width=\linewidth]{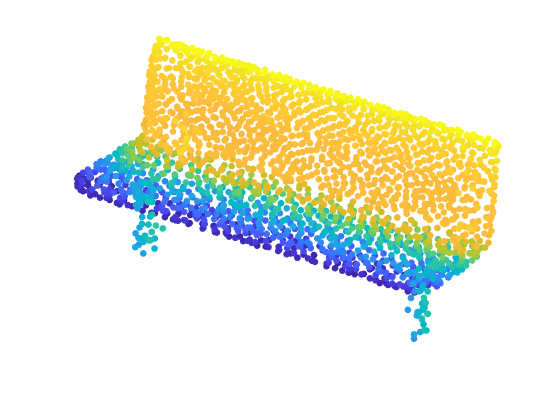}
    \end{minipage}
    &
    \begin{minipage}{.12\linewidth}
      \includegraphics[trim=2cm 0.5cm 2cm 0.5cm,clip,width=\linewidth]{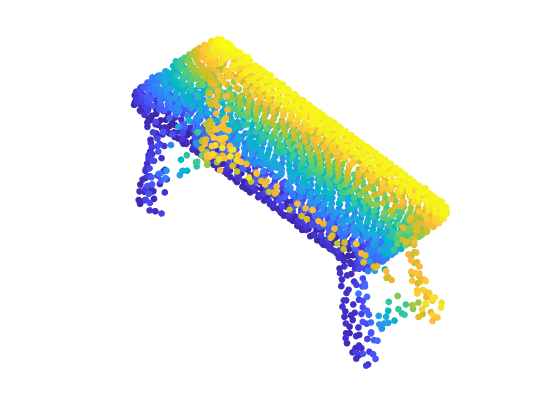}
    \end{minipage}
    &
    \begin{minipage}{.12\linewidth}
      \includegraphics[trim=2cm 0.5cm 2cm 0.5cm,clip,width=\linewidth]{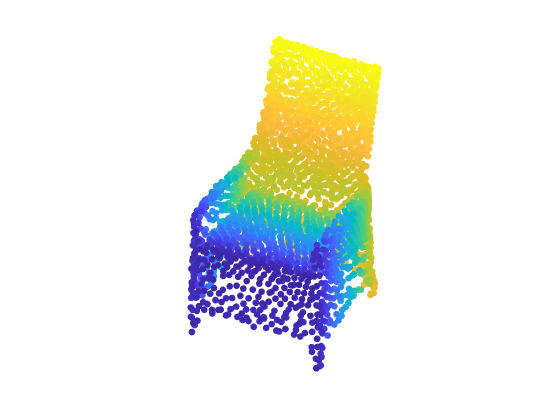}
    \end{minipage}
    & 
    \begin{minipage}{.12\linewidth}
      \includegraphics[trim=2cm 0.5cm 2cm 0.5cm,clip,width=\linewidth]{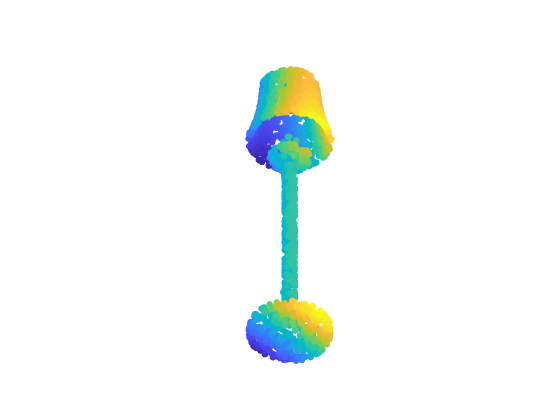}
    \end{minipage}
    &
    \begin{minipage}{.12\linewidth}
      \includegraphics[trim=2cm 0.5cm 2cm 0.5cm,clip,width=\linewidth]{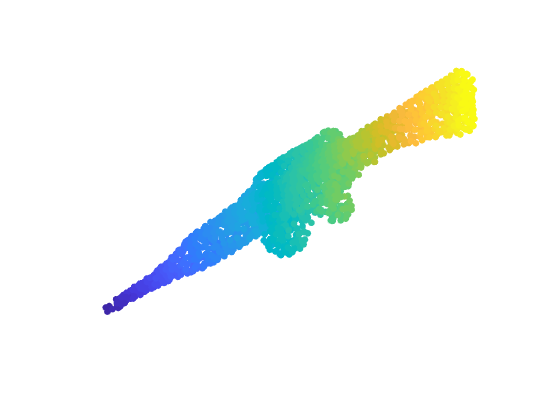}
    \end{minipage}
    \\ 
  \end{tabular}
  \end{center}
\captionof{figure}{Examples of point cloud reconstruction results on ShapeNetCore13. (a) is the ground truth and (b) is our result.}
\label{fig:fig3}
\vspace{-0.1cm}
\end{table*}

\begin{table}[!ht]
\begin{center}
\begin{tabular}{l>{\centering}m{15mm}>{\centering\arraybackslash}m{15mm}}
\toprule
Method & ShapeNet \newline Core13 & ShapeNet \newline Corev2 \\
\midrule
Oracle & 0.85 & 3.10\\
L-GAN~\cite{lgan} & - & 7.07\\
Points baseline & 1.61 & -\\
AtlasNet-125~\cite{atlasnet} & 1.51 & \textbf{5.66}\\
PointFlow~\cite{pointflow} & - & 7.54\\
3D-PointCapsNet~\cite{3dpointcapsulenet} & 1.46 & -\\
DF-Net~\cite{dfnet} & - & 6.17\\
\midrule
Ours & \textbf{1.39} & 5.78\\
\bottomrule
\end{tabular}
\end{center}
\caption{Results of point cloud reconstruction on ShapeNetCore13 and ShapeNetCorev2. The chamfer distance is multiplied by $10^{3}$ for ShapeNetCore13 and $10^{4}$ for ShapeNetCorev2.}
\label{tab:tab1}
\vspace{-0.1cm}
\end{table}

\begin{definition}
The problem is redefined as finding the function $\Psi'$, where $\Psi'(P)$ forms an identity function $I_{P}$.
\end{definition}

Furthermore, we expand 2-dimensional point set $U'$ to an $n$-dimensional features that contains more semantic information than simple 2D coordinates. 
This guarantees more informative $u$ to generate $p'$. 

Another noticeable difference between our method and previous methods is that our method produces the output in a progressive manner. Because progressive seeds with multiple resolutions are generated through the SGMs, we can combine coarse semantic information and fine detailed information by fusing the feature hierarchy from each seed. Similar to FCN~\cite{fcn}, we implement feature fusion by a $1\times1$ Conv - BN - ReLU layer sequence and a bilinear interpolation operation. 

In summary, there are three factors that have led to significant differences between previous methods~\cite{foldingnet, atlasnet, 3dpointcapsulenet}.
\begin{enumerate}
    \item A discrete set $U' \subset \mathbb{R}^{2}$
    \item Parameterization function $\Psi$
    \item Combination of $U'$ and a codeword vector $\theta$
\end{enumerate}
Our framework uses Input-dependent high-dimensional seeds that facilitate the reconstruction, while existing methods used arbitrary 2D coordinates. Since we adopted transposed 2D Conv in the process of generating input-dependent seeds for simplicity and efficiency, our $\Psi$ is defined as the transposed 2D Conv, MLP (equivalent to $1\times1$ Conv), and bilinear interpolation. Naturally, the concatenation of $\theta$, $U'$, and the corresponding coarse feature becomes the input of $\Psi$.

In addition, we can control the number of output point clouds by adjusting the parameters that make up the decoder, as in existing methods. We show an example of this in the point cloud upsampling task, please refer to Supplementary Material Section 2.

\section{Experiments}
We conducted extensive experiments to demonstrate the effectiveness of our method. First, we compared the performance of our method against existing methods for various 3D tasks such as point cloud reconstruction, unsupervised classification and point cloud completion. Then, we experimentally proved the analysis provided in Section~\ref{sec:sec_3_3}.

\begin{table}[!ht]
\begin{center}
\begin{tabular}{lcc}
\toprule
Method & Supervised & Accuracy (\%) \\
\midrule
PointNet~\cite{pointnet} & Yes & 89.2\\
PointNet++~\cite{pointnet++} & Yes & 90.7\\
PointCNN~\cite{pointcnn} & Yes & 92.2\\
DGCNN~\cite{dgcnn} & Yes & 92.2\\
\midrule
L-GAN~\cite{lgan} & No &  85.7\\
L-GAN~\cite{lgan} (MN40)& No & 87.3\\
FoldingNet~\cite{foldingnet} & No & 88.4\\
FoldingNet~\cite{foldingnet} (MN40) & No &  84.4\\
3D-PointCapsNet~\cite{3dpointcapsulenet} & No & 88.9\\
PointGrow~\cite{pointgrow} & No & 85.8\\
PointFlow~\cite{pointflow} & No & 86.8\\
MRTNet-VAE~\cite{mrtnetvae} & No & 86.4 \\
PCGAN~\cite{pcgan} & No & 87.8\\
SA-Net-cls~\cite{sanet} & No & 90.6\\
\midrule
Ours & No & \textbf{90.9}\\
\bottomrule
\end{tabular}
\end{center}
\caption{The results of unsupervised classification on ModelNet40.}
\label{tab:tab3}
\vspace{-0.1cm}
\end{table}

\subsection{Implementation}
Our method is evaluated on four datasets: ShapeNetCore13~\cite{shapenet}, ShapeNetCorev2~\cite{shapenet}, Completion3D~\cite{topnet}, and ModelNet40~\cite{modelnet} datasets. We implemented our network using the PyTorch~\cite{pytorch} framework. The network was trained using an ADAM optimizer with betas (0.9, 0.999) and weight decay 1e-6. We used an initial learning rate of 5e-5 for ModelNet40, and 1e-4 for others. The batch size was set to 32. The chamfer distance (CD) was used as the evaluation metric of point cloud reconstruction and completion tasks.

\subsection{Comparative Study}
\label{sec:sec_4_2}

\begin{table*}[!ht]
\begin{center}
\begin{adjustbox}{max width=\textwidth}
\begin{tabular}{lccccccccc}
\toprule
Method & Overall & Airplane & Cabinet & Car & Chair & Lamp & Sofa & Table & Watercraft\\
\midrule

FoldingNet~\cite{foldingnet} & 19.07 & 12.83 & 23.01 & 14.88 & 25.69 & 21.79 & 21.31 & 20.71 & 11.51\\

PCN~\cite{PCN} & 18.22 & 9.79 & 22.70 & 12.43 & 25.14 & 22.72 & 20.26 & 20.27 & 11.73\\

PointSetVoting~\cite{pointsetvoting} & 18.18 & 6.88 & 21.18 & 15.78 & 22.54 & 18.78 & 28.39 & 19.96 & 11.16\\

AtlasNet~\cite{atlasnet} & 17.77 & 10.36 & 23.40 & 13.40 & 24.16 & 20.24 & 20.82 & 17.52 & 11.62\\

PointNetFCAE & 16.88 & 10.30 & 19.06 & 11.82 & 24.68 & 20.30 & 20.09 & 17.57 & 10.50\\

TopNet~\cite{topnet} & 14.25 & 7.32 & 18.77 & 12.88 & 19.82 & 14.60 & 16.29 & 14.89 & 8.82\\

\midrule

SoftPoolNet~\cite{softpoolnet} & 11.90 & 4.89 & 18.86 & 10.17 & 15.22 & 12.34 & 14.87 & 11.84 & 6.48\\

SA-Net~\cite{sanet} & 11.22 & 5.27 & 14.45 & 7.78 & 13.67 & 13.53 & 14.22 & 11.75 & 8.84\\

GRNet~\cite{grnet} & 10.64 & 6.13 & 16.90 &8.27 & 12.23 & 10.22 & 14.93 & 10.08 & 5.86\\

PMP-Net~\cite{pmpnet} & 9.23 & 3.99 & 14.70 & 8.55 & 10.21 & 9.27 & 12.43 & 8.51 & 5.77\\

\midrule

Ours & 13.29 & 7.19 & 19.54 & 10.5 & 18.53 & 14.16 & 16.8 & 11.34 & 7.69\\
\bottomrule
\end{tabular}
\end{adjustbox}
\end{center}
\caption{Results of point cloud completion on Completion3D. The chamfer distance is multiplied by $10^{3}$.}
\label{tab:tab2}
\vspace{-0.1cm}
\end{table*}

\begin{table*}[!ht]
  \begin{center}
  \centering
  \begin{tabular}{ ccccccc }
    (a) &
    \begin{minipage}{.12\linewidth}
      \includegraphics[trim=2cm 0.5cm 2cm 0.5cm,clip,width=\linewidth]{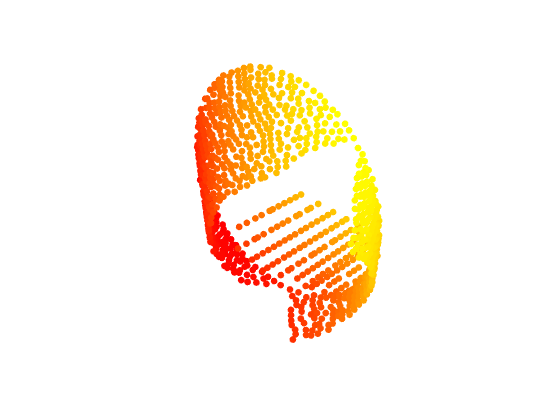}
    \end{minipage}
    & 
    \begin{minipage}{.12\linewidth}
      \includegraphics[trim=2cm 0.5cm 2cm 0.5cm,clip,width=\linewidth]{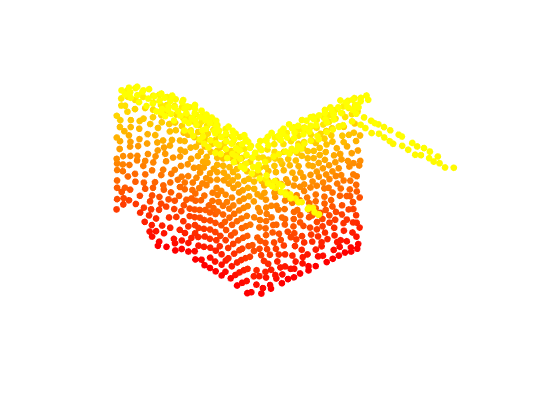}
    \end{minipage}
    &
    \begin{minipage}{.12\linewidth}
      \includegraphics[trim=2cm 0.5cm 2cm 0.5cm,clip,width=\linewidth]{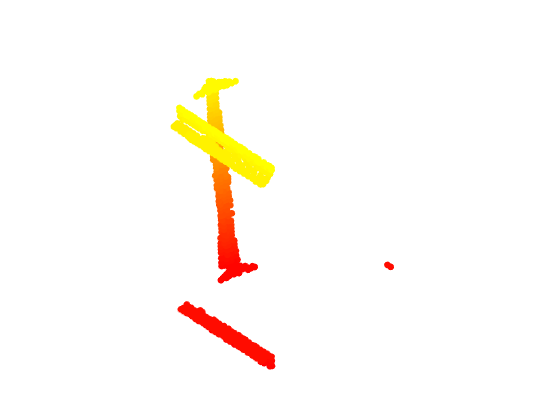}
    \end{minipage}
    &
    \begin{minipage}{.12\linewidth}
      \includegraphics[trim=2cm 0.5cm 2cm 0.5cm,clip,width=\linewidth]{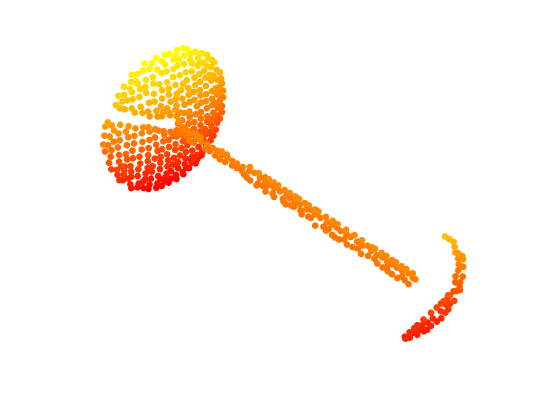}
    \end{minipage}
    & 
    \begin{minipage}{.12\linewidth}
      \includegraphics[trim=3cm 2cm 3cm 2cm,clip,width=\linewidth]{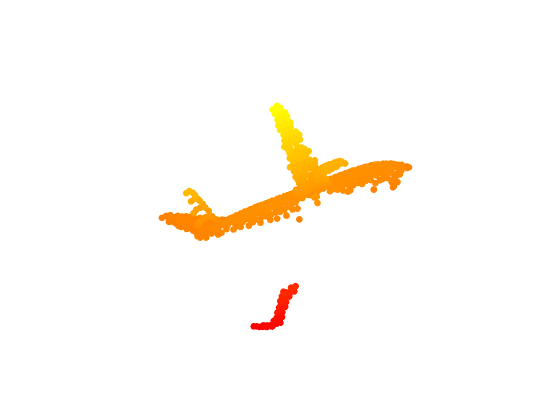}
    \end{minipage}
    &
    \begin{minipage}{.12\linewidth}
      \includegraphics[trim=2cm 1.5cm 2cm 0.5cm,clip,width=\linewidth]{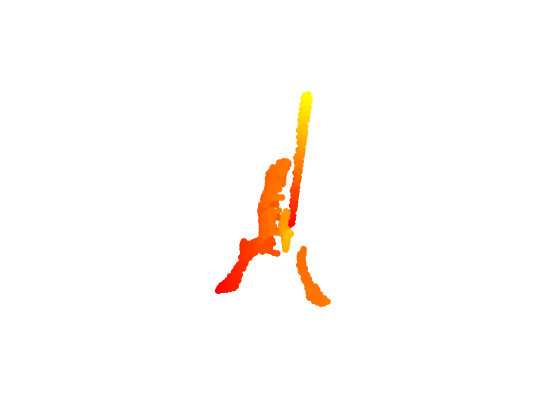}
    \end{minipage}
    \\ 
    (b) &
    \begin{minipage}{.12\linewidth}
      \includegraphics[trim=2cm 0.5cm 2cm 0.5cm,clip,width=\linewidth]{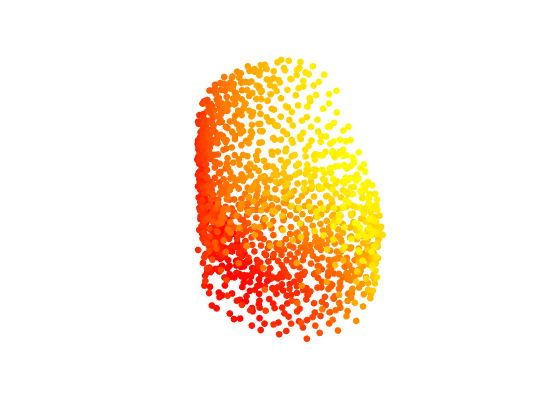}
    \end{minipage}
    & 
    \begin{minipage}{.12\linewidth}
      \includegraphics[trim=2cm 0.5cm 2cm 0.5cm,clip,width=\linewidth]{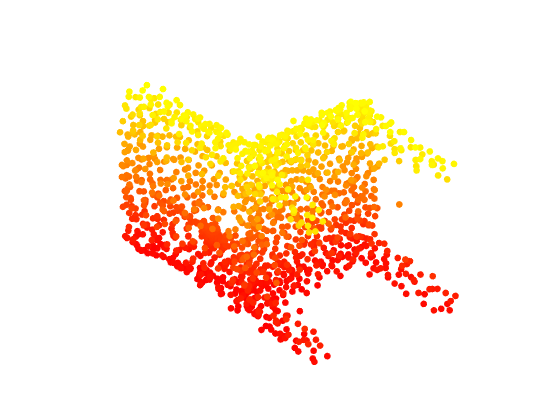}
    \end{minipage}
    &
    \begin{minipage}{.12\linewidth}
      \includegraphics[trim=2cm 0.5cm 2cm 0.5cm,clip,width=\linewidth]{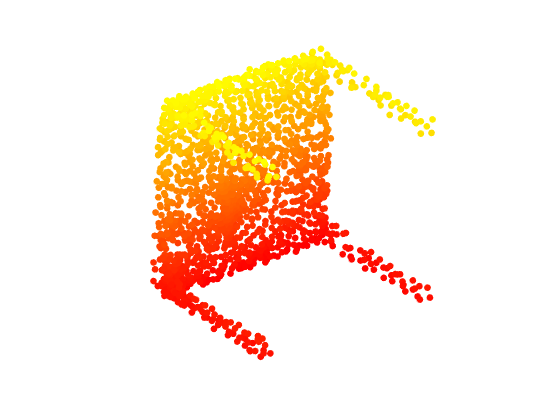}
    \end{minipage}
    &
    \begin{minipage}{.12\linewidth}
      \includegraphics[trim=2cm 0.5cm 2cm 0.5cm,clip,width=\linewidth]{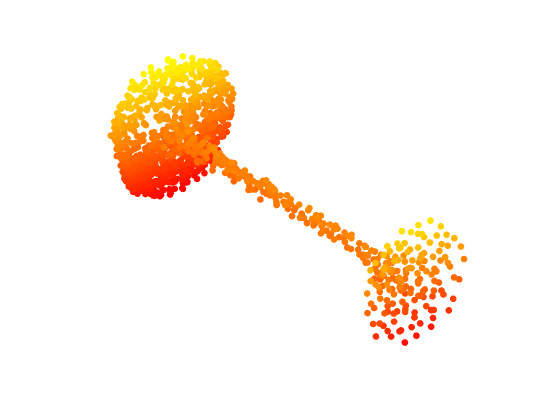}
    \end{minipage}
    & 
    \begin{minipage}{.12\linewidth}
      \includegraphics[trim=3cm 2cm 3cm 2cm,clip,width=\linewidth]{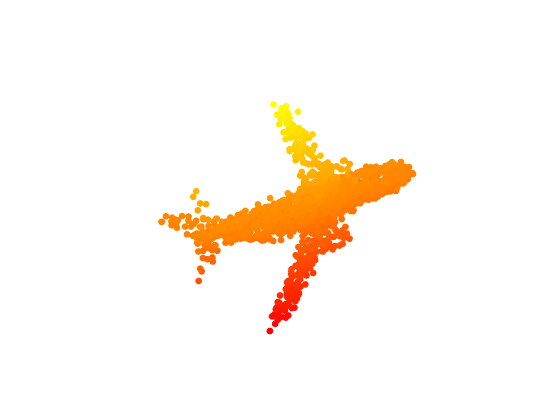}
    \end{minipage}
    &
    \begin{minipage}{.12\linewidth}
      \includegraphics[trim=2cm 1.5cm 2cm 0.5cm,clip,width=\linewidth]{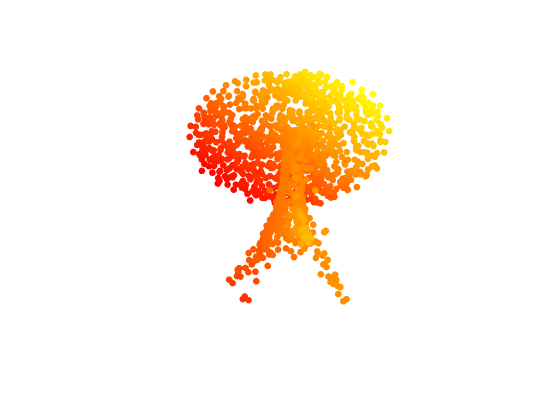}
    \end{minipage}
    \\ 
    (c) &
    \begin{minipage}{.12\linewidth}
      \includegraphics[trim=2cm 0.5cm 2cm 0.5cm,clip,width=\linewidth]{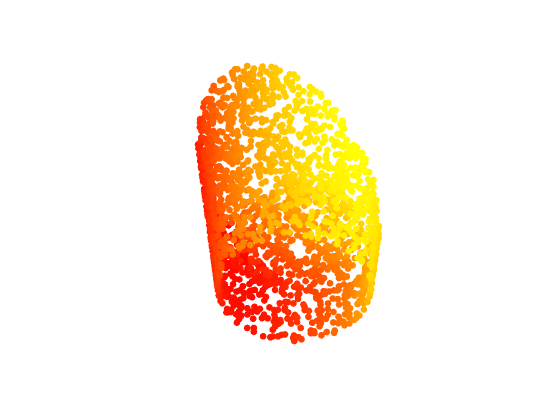}
    \end{minipage}
    & 
    \begin{minipage}{.12\linewidth}
      \includegraphics[trim=2cm 0.5cm 2cm 0.5cm,clip,width=\linewidth]{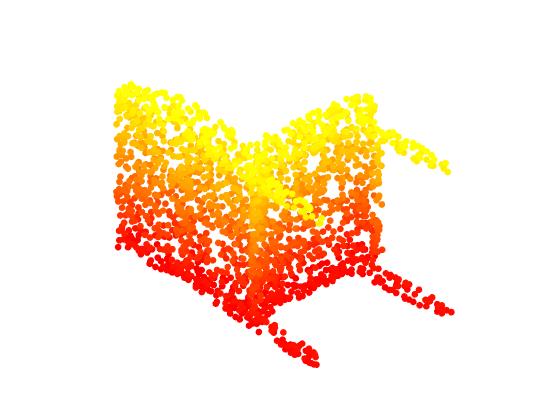}
    \end{minipage}
    &
    \begin{minipage}{.12\linewidth}
      \includegraphics[trim=2cm 0.5cm 2cm 0.5cm,clip,width=\linewidth]{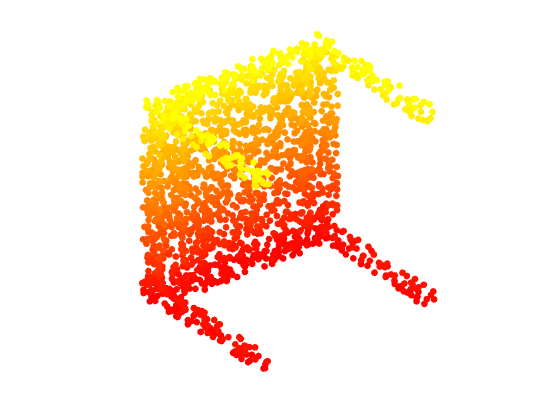}
    \end{minipage}
    &
    \begin{minipage}{.12\linewidth}
      \includegraphics[trim=2cm 0.5cm 2cm 0.5cm,clip,width=\linewidth]{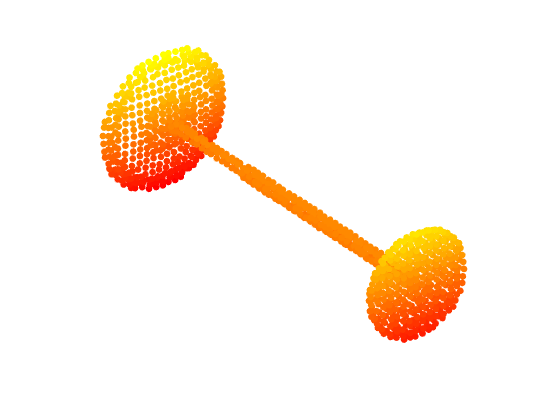}
    \end{minipage}
    & 
    \begin{minipage}{.12\linewidth}
      \includegraphics[trim=3cm 2cm 3cm 2cm,clip,width=\linewidth]{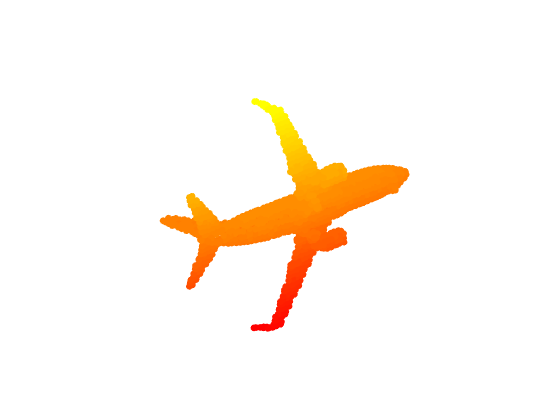}
    \end{minipage}
    &
    \begin{minipage}{.12\linewidth}
      \includegraphics[trim=2cm 1.5cm 2cm 0.5cm,clip,width=\linewidth]{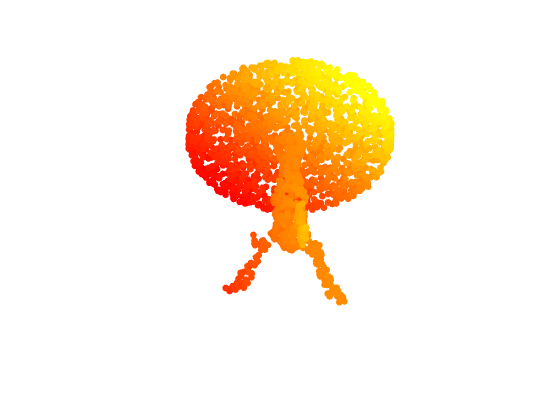}
    \end{minipage}
    \\ 
  \end{tabular}
  \end{center}
\captionof{figure}{Examples of point cloud completion results on Completion3D. (a) is the input, (b) is our result, and (c) is the ground truth.}
\label{fig:fig4}
\vspace{-0.2cm}
\end{table*}

\textbf{Point cloud reconstruction.}
We performed point cloud reconstruction on ShapeNetCore13 and ShapeNetCorev2. We followed the dataset settings of AtlasNet~\cite{atlasnet} for both datasets. The results in Table~\ref{tab:tab1} clearly show that our model achieves performance that is comparable to the state-of-the-art methods. Oracle randomly samples a point cloud from the ground truth shape, and therefore constitutes an upper bound of the performance. Points baseline is a network of MLPs presented with detailed structures in AtlasNet.

It should be noted that AtlasNet and 3D-PointCapsNet utilize 125 and 32 patches for finer reconstruction, respectively. Our method uses only a single patch but outperforms the two methods on ShapeNetCore13, and achieves comparable performance to AtlasNet on ShapeNetCorev2. We further present performance that outperforms even AtlasNet in Section~\ref{sec:sec_4_3}. Figure~\ref{fig:fig3} visually shows the reconstruction results of ShapeNetCore13.

\textbf{Unsupervised classification.}
To evaluate the efficiency of the feature representation of our method, we performed unsupervised classification on ModelNet40. In accordance with the settings of the previous studies~\cite{lgan, foldingnet, 3dpointcapsulenet}, we first performed the reconstruction on ModelNet40 and then extracted the codeword vector. Thereafter, we normalized the codeword vector and used it to train the linear support vector machine (SVM) classifier. Table~\ref{tab:tab3} indicates that our method achieves reasonable accuracy among unsupervised learning methods. We note that our method not only outperforms the unsupervised methods, but also surpasses the supervised version of the same encoder. 

\textbf{Point cloud completion.}
To demonstrate the effectiveness of our decoder architecture, we train PSG-Net for supervised learning of point cloud completion.
Because the experimental settings of existing studies on point cloud completion have not been unified, we participated in the Completion3D benchmark\footnote{https://completion3d.stanford.edu} to evaluate our model for the point cloud completion task. 
We utilized the PCN~\cite{PCN} encoder, which is mainly used in the completion task for comparison focused on the decoder architecture.

We added our result to the current leaderboard and present it in Table~\ref{tab:tab2}. We excluded the result of no method being presented. The result shows that the performance of our method is comparable to that of others for the point cloud completion task, even though the structure is aimed at reconstruction. The methods which show better performance than ours, are not suitable for direct comparison with ours because they use the codeword vector and other techniques with much powerful encoders for supervised point cloud completion. 
However, our method is aimed at unsupervised learning and uses only global representation with PCN encoder. Thus, practically, our method can be considered to have achieved the best performance in counterpart methods, and improvements can be expected depending on the type of encoder. Because no ground truth for the test set is publicly available, we visualize the completion results of the validation set in Figure~\ref{fig:fig4}.

\begin{table}[t]
\begin{center}
\begin{tabular}{cc}
\toprule
Method & ShapeNetCorev2 \\
\midrule
Ours (512-dim) & 5.78\\
Ours (1024-dim) & \textbf{5.15}\\
\bottomrule
\end{tabular}
\end{center}
\caption{Results of point cloud reconstruction on ShapeNetCorev2. The chamfer distance is multiplied by $10^{4}$ for ShapeNetCorev2.}
\label{tab:tab5}
\end{table}

\begin{table}[t]
\begin{center}
\begin{tabular}{cc}
\toprule
Method & \# Parameters (M) \\
\midrule
AtlasNet-25~\cite{atlasnet} & 44.8\\
AtlasNet-125~\cite{atlasnet} & 219.3\\
3D-PointCapsNet~\cite{3dpointcapsulenet} & 69.5\\
\midrule
Ours & \textbf{7.6}\\
\bottomrule
\end{tabular}
\end{center}
\caption{The number of parameters on various networks.}
\label{tab:tab6}
\vspace{-0.2cm}
\end{table}

\subsection{Analytical Study}
\label{sec:sec_4_3}
We explained in Section~\ref{sec:sec_3_3} that our method is superior to existing methods. In this section, we demonstrate this experimentally in various ways.

\textbf{Effect of different codeword lengths.}
Because the existing methods used 512-dimensional or 1024-dimensional codeword vectors, we conducted experiments based on 512-dimensional codeword vectors. However, to show the effect of using different codeword lengths, we performed point cloud reconstruction on ShapeNetCorev2 with 1024-dimensional codeword vector. The results in Table~\ref{tab:tab5} clearly demonstrate that our method performs well with 1024-dimensional codeword vector.

\textbf{Computational efficiency.}
To demonstrate the efficiency of our method, we compare the number of parameters of our method with those of the existing methods. Table~\ref{tab:tab6} indicates that our network requires far fewer parameters than existing methods. As mentioned in Section~\ref{sec:sec1}, the number of parameters increases linearly with the number of patches, which is shown by the examples of AtlasNet-25 and AtlasNet-125~\cite{atlasnet}. This observation emphasizes the computational efficiency of our method.

\begin{table}[!hbt]
\begin{center}
\begin{tabular}{lccc}
\toprule
Task & Method & Supervised & Result\\
\midrule
rec & AtlasNet-125~\cite{atlasnet} & No & 5.66\\
rec & Ours + AtlasNet-32~\cite{atlasnet} & No & \textbf{5.31}\\
\midrule
cls & PointGLR~\cite{globallocal} & Yes & 91.69\\
cls & PointGLR~\cite{globallocal} & No & 92.22\\
cls & PointGLR~\cite{globallocal} & Hybrid & 92.42\\
cls & Ours + PointGLR~\cite{globallocal} & No & \textbf{92.59}\\
\bottomrule
\end{tabular}
\end{center}
\caption{The results of combination with other methods. Rec and cls indicate reconstruction task and unsupervised classification task, respectively. In result, reconstruction task uses the chamfer distance multiplied by $10^{4}$ as an evaluation metric, and classification task uses the accuracy (\%) as an evaluation metric.} 
\vspace{-0.2cm}
\label{tab:tab7}
\end{table}

\begin{table*}[!ht]
\begin{center}
\begin{adjustbox}{max width=\textwidth}
\begin{tabular}{cccccccc}
\toprule
Method & Decoder $A$ & Decoder $B$ & Decoder $C_2$ & Decoder $C_{32}$ & + 1 SFPM & + 2 SFPMs & + 3 SFPMs\\
\midrule
CD & 7.43 & 7.96 & 7.34 & 6.84 & 6.63 & 6.40 & \textbf{6.38}\\
\bottomrule
\end{tabular}
\end{adjustbox}
\end{center}
\caption{Results of point cloud reconstruction on ShapeNetCorev2 with various decoders for analysis. The chamfer distance is multiplied by $10^{4}$.}
\label{tab:tab4}
\vspace{-0.4cm}
\end{table*}

\begin{table*}[!ht]
  \begin{center}
  \centering
  \begin{tabular}{ cccc }
    
    \begin{minipage}{.21\linewidth}
      \includegraphics[trim=2cm 2cm 2cm 2cm,clip,width=\linewidth]{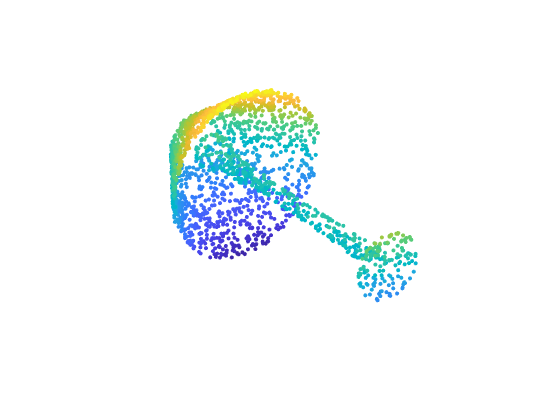}
    \end{minipage}
    &
    \begin{minipage}{.21\linewidth}
      \includegraphics[trim=2cm 2cm 2cm 2cm,clip,width=\linewidth]{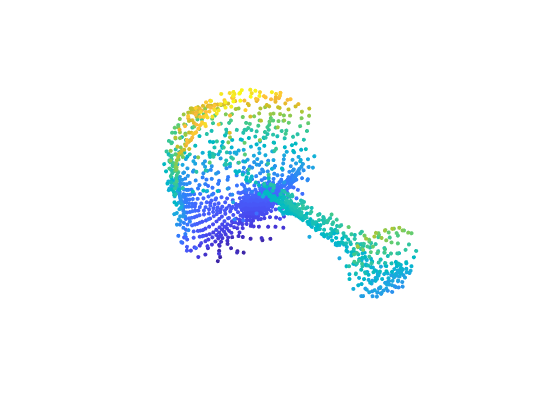}
    \end{minipage}
    & 
    \begin{minipage}{.21\linewidth}
      \includegraphics[trim=2cm 2cm 2cm 2cm,clip,width=\linewidth]{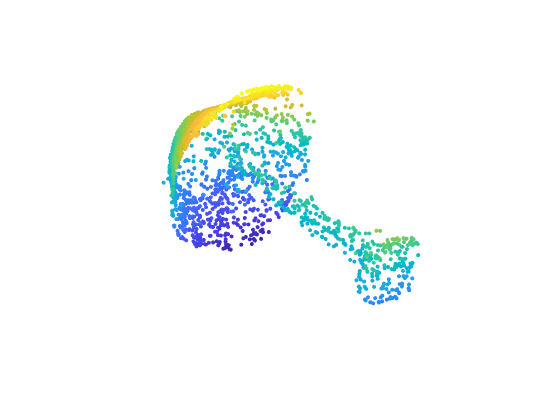}
    \end{minipage}
    &
    \begin{minipage}{.21\linewidth}
      \includegraphics[trim=2cm 2cm 2cm 2cm,clip,width=\linewidth]{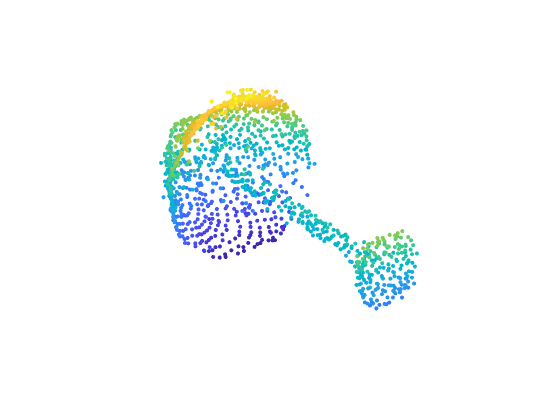}
    \end{minipage}
    \\ 
    \begin{minipage}{.21\linewidth}
      \includegraphics[trim=0cm 0cm 0cm 0cm,clip,width=\linewidth]{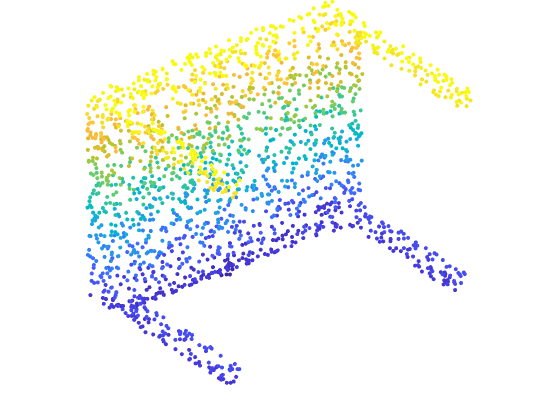}
    \end{minipage}
    &
    \begin{minipage}{.21\linewidth}
      \includegraphics[trim=0cm 0cm 0cm 0cm,clip,width=\linewidth]{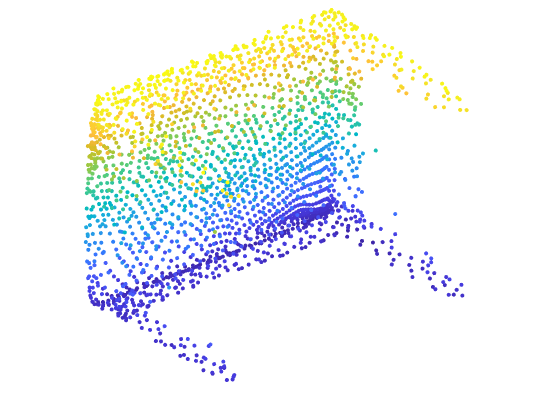}
    \end{minipage}
    & 
    \begin{minipage}{.21\linewidth}
      \includegraphics[trim=0cm 0cm 0cm 0cm,clip,width=\linewidth]{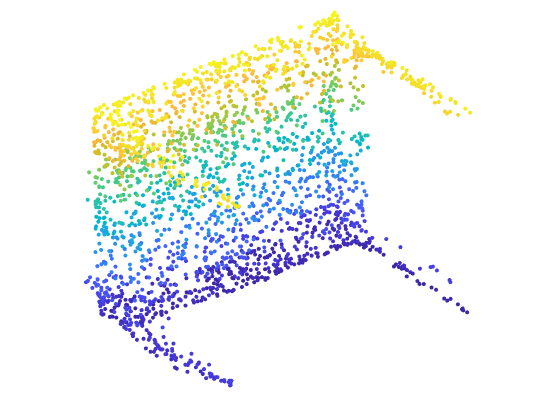}
    \end{minipage}
    &
    \begin{minipage}{.21\linewidth}
      \includegraphics[trim=0cm 0cm 0cm 0cm,clip,width=\linewidth]{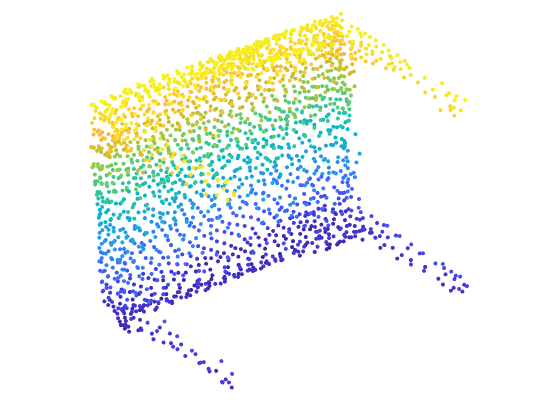}
    \end{minipage}
    \\ 
    \\
    Input & Decoder $A$ & Decoder $B$ & Decoder $C_{32}$
  \end{tabular}
  \end{center}
  \vspace{0.1cm}
\captionof{figure}{Examples of point cloud reconstruction on ShapeNetCorev2 with various decoders for analysis.}
\label{fig:fig5}
\vspace{-0.1cm}
\end{table*}

\begin{table*}[!ht]
  \begin{center}
  \centering
  \begin{tabular}{ ccc }
    \begin{minipage}{.25\linewidth}
      \includegraphics[trim=0cm 0cm 0cm 0cm,clip,width=\linewidth]{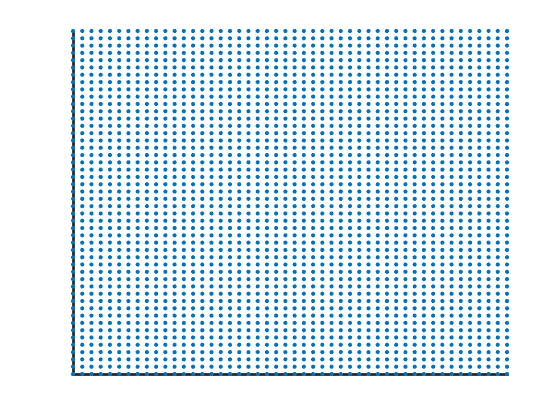}
    \end{minipage}
    &
    \begin{minipage}{.25\linewidth}
      \includegraphics[trim=0cm 0cm 0cm 0cm,clip,width=\linewidth]{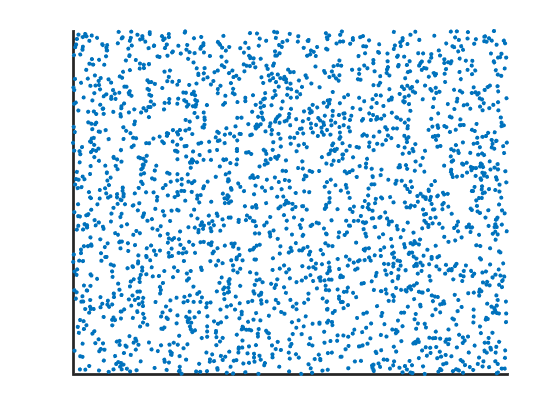}
    \end{minipage}
    &
    \begin{minipage}{.25\linewidth}
      \includegraphics[trim=0cm 0cm 0cm 0cm,clip,width=\linewidth]{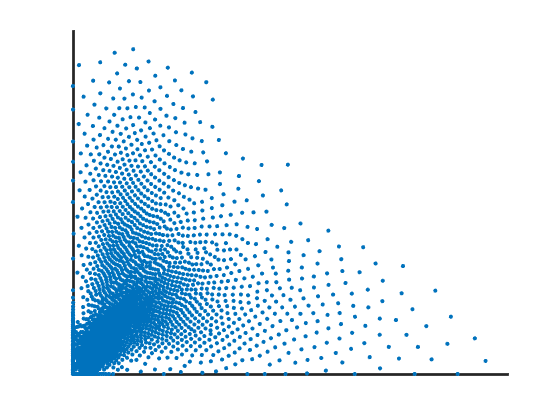}
    \end{minipage}
    \\ 
     Decoder $A$  & Decoder $B$  & Decoder $C_2$
  \end{tabular}
  \end{center}
\captionof{figure}{Coordinates of the sampled points used in Decoders $A$, $B$, and $C_2$. Decoder $A$ used grid points, Decoder $B$ used random points, and Decoder $C_2$ used points constructed from the codeword vector.}
\label{fig:fig2}
\vspace{-0.1cm}
\end{table*}

\textbf{Combination with other methods.}
We conducted experiments that produced higher performance, even state-of-the-art, through combinations with existing methods to demonstrate the applicability of our method. For the reconstruction on ShapeNetCorev2, we applied our method to AtlasNet by adding the SGM to the structure of AtlasNet with 32 patches.

For the unsupervised classification, since the results in Table~\ref{tab:tab3} show the results of training each proposed network by normal reconstruction training with a chamfer distance, studies with higher performance may exist as new unsupervised learning methods are applied. For example, \cite{globallocal} focused on new loss functions which help unsupervised learning, rather than the network architecture itself, and achieved state-of-the-art performance on the unsupervised classification task. Although \cite{globallocal} presented higher performance, we did not compare our method with \cite{globallocal} in Table~\ref{tab:tab3} because we considered that their improvement is orthogonal to that of our work. Thus, we further conducted an experiment that combines our decoder with \cite{globallocal}.

The results in Table~\ref{tab:tab7} show that our method not only achieves the state-of-the-art, but is capable of combining with other studies.

\textbf{Decoder ablation.}
We constructed various models suitable for analysis and conducted point cloud reconstruction on the ShapeNetCorev2. Our analysis was focused on the contribution of different $U'$s and the progressive seed generation to the performance improvement.

We first compared the performance of various $U'$ settings. For fair comparison, we used the same encoder based on PointNet~\cite{pointnet} and the same decoder based on FoldingNet~\cite{foldingnet}, which consists of two MLP - BN - ReLU sequences. We built the following four decoders which differ only in $U'$: \textbf{Decoder $\bm{A}$} that sets $U'$ as fixed 2D grid points; \textbf{Decoder $\bm{B}$} that sets $U'$ as 2D points sampled from the uniform distribution $\mathcal{U}(0,1)$; \textbf{Decoder $\bm{C_2}$} that generates $U'$ from $\theta$; \textbf{Decoder $\bm{C_{32}}$} that generates $U'$ as 32-dimensional features generated from $\theta$ (Please refer to Supplementary Material Section 3).

The results in Table~\ref{tab:tab4} and Figure~\ref{fig:fig5} demonstrate that our method performs point cloud reconstruction more effectively than existing methods. Interestingly, we observe that Decoder $A$ with $U'$ set to a fixed grid has a higher performance than Decoder $B$ with $U'$ set to a uniform distribution. It can also be seen that Decoder $C_2$ performs better than both Decoders $A$ and $B$, and that the performance is even better when high-dimensional feature vectors are used instead of 2D points (Decoder $C_{32}$). These results are consistent with the analysis in Section~\ref{sec:sec_3_3}. To form $\Psi$ as an identity function in Def~\ref{def:def3}, it is best to fix the independent variable $U'$ to a constant value. Therefore, Decoder $A$ with $U'$ set to a fixed 2D grid can learn a function better than Decoder $B$ with $U'$ set to a random variable. In addition, because $U'$ is a function of $P$, Decoder $C_2$ achieves higher performance than Decoders $A$ and $B$. Decoder $C_{32}$ achieves the highest performance because $U'$ contains the most information about $P$. Figure~\ref{fig:fig2} shows the coordinates of the 2D points used in Decoders $A$ and $B$ and the coordinates of 2D points learned in Decoder $C_2$. We observe that the Decoders $A$ and $B$ sample grid points and completely random points on a rectangular plane, whereas the Decoder $C_2$ samples points on a uniquely constructed plane.

We also prove the effectiveness of the progressive approach by adding more SFPMs to Decoder $C_{32}$ and observe the performance change in point cloud reconstruction. As shown in Table~\ref{tab:tab4}, the performance is gradually improved as 1, 2, and 3 SFPMs are added.


\section{Conclusion}
In this paper, we propose a novel framework that generates the input-dependent point-wise features as seed for reconstruction-based learning of point cloud. 
To this end, we constructed efficient yet powerful PSG-Net with two modules, namely SGM and SFPM, focusing on the concept of the seed. 
PSG-Net achieves reasonable performance in point cloud reconstruction, unsupervised classification, and point cloud completion. Given that we focus on the fundamental concept rather than the technical concept of point cloud reconstruction and the construction of PSG-Net, this study can be used with other methods. Considering the application of PSG-Net, we believe that our study will contribute to the understanding and application of point clouds.

\section*{Acknowledgements}
This research was supported by Naver Labs Corporation.

{\small
\bibliographystyle{ieee_fullname}
\bibliography{egbib}
}

\end{document}